\documentclass[conference]{IEEEtran}
\IEEEoverridecommandlockouts
\usepackage{cite}
\usepackage{amsmath,amssymb,amsfonts}
\usepackage{algorithmic}
\usepackage{graphicx}
\usepackage{paralist}
\usepackage{epstopdf}
\usepackage{textcomp}
\usepackage{xcolor}
\usepackage{algorithm}
\usepackage{tabularray}
\usepackage{bm}
\UseTblrLibrary{diagbox}

\def\BibTeX{{\rm B\kern-.05em{\sc i\kern-.025em b}\kern-.08em
    T\kern-.1667em\lower.7ex\hbox{E}\kern-.125emX}}
\begin{document}

\title{RaftFed: A Lightweight Federated Learning Framework for Vehicular Crowd Intelligence\\

}

 \author{
     \IEEEauthorblockN{Changan Yang\IEEEauthorrefmark{1},
         Yaxing Chen\IEEEauthorrefmark{2},
         Yao Zhang\IEEEauthorrefmark{2}, 
         Helei Cui\IEEEauthorrefmark{2}, 
         Zhiwen Yu\IEEEauthorrefmark{2},
         Bin Guo\IEEEauthorrefmark{2},
         Zheng Yan\IEEEauthorrefmark{3},
         Zijiang Yang \IEEEauthorrefmark{4}}
     \vspace{4pt}
     \IEEEauthorblockA{\IEEEauthorrefmark{1}School of Software, Northwestern Polytechnical University, Xi'an China}
     \IEEEauthorblockA{\IEEEauthorrefmark{2}School of Computer Science, Northwestern Polytechnical University, Xi'an, China}
     \IEEEauthorblockA{\IEEEauthorrefmark{3}School of Cyber Engineering, Xidian University, Xi'an, China}
     \IEEEauthorblockA{\IEEEauthorrefmark{4}Turing Research Center for Interdisciplinary Information Science, Xi'an Jiaotong University, Xi'an, China}
     \IEEEauthorblockA{Email:\IEEEauthorrefmark{1}yca.npu@gmail.com,\IEEEauthorrefmark{2}\{yxchen, yaozh.g, chl, zhiwenyu,  guob\}@nwpu.edu.cn,\IEEEauthorrefmark{3}zyan@xidian.edu.cn, \IEEEauthorrefmark{4}zijiang@xjtu.edu.cn}
 }
\maketitle

\begin{abstract}
Vehicular crowd intelligence (VCI) is an emerging research field. Facilitated by state-of-the-art vehicular ad-hoc networks and artificial intelligence, various VCI applications come to place, e.g., collaborative sensing, positioning, and mapping. The collaborative property of VCI applications generally requires data to be shared among participants, thus forming network-wide intelligence. How to fulfill this process without compromising data privacy remains a challenging issue. Although federated learning (FL) is a promising tool to solve the problem, adapting conventional FL frameworks to VCI is nontrivial. First, the centralized model aggregation is unreliable in VCI because of the existence of stragglers with unfavorable channel conditions. Second, existing FL schemes are vulnerable to Non-IID data, which is intensified by the data heterogeneity in VCI. This paper proposes a novel federated learning framework called RaftFed to facilitate privacy-preserving VCI. The experimental results show that RaftFed performs better than baselines regarding communication overhead, model accuracy, and model convergence.
\end{abstract}

\begin{IEEEkeywords}
Vehicular Ad-hoc Networks, Federated Learning, Crowd Intelligence, Dynamic Clustering
\end{IEEEkeywords}

\section{Introduction}
\begin{figure}[!h]
\centerline{\includegraphics[scale=0.06]{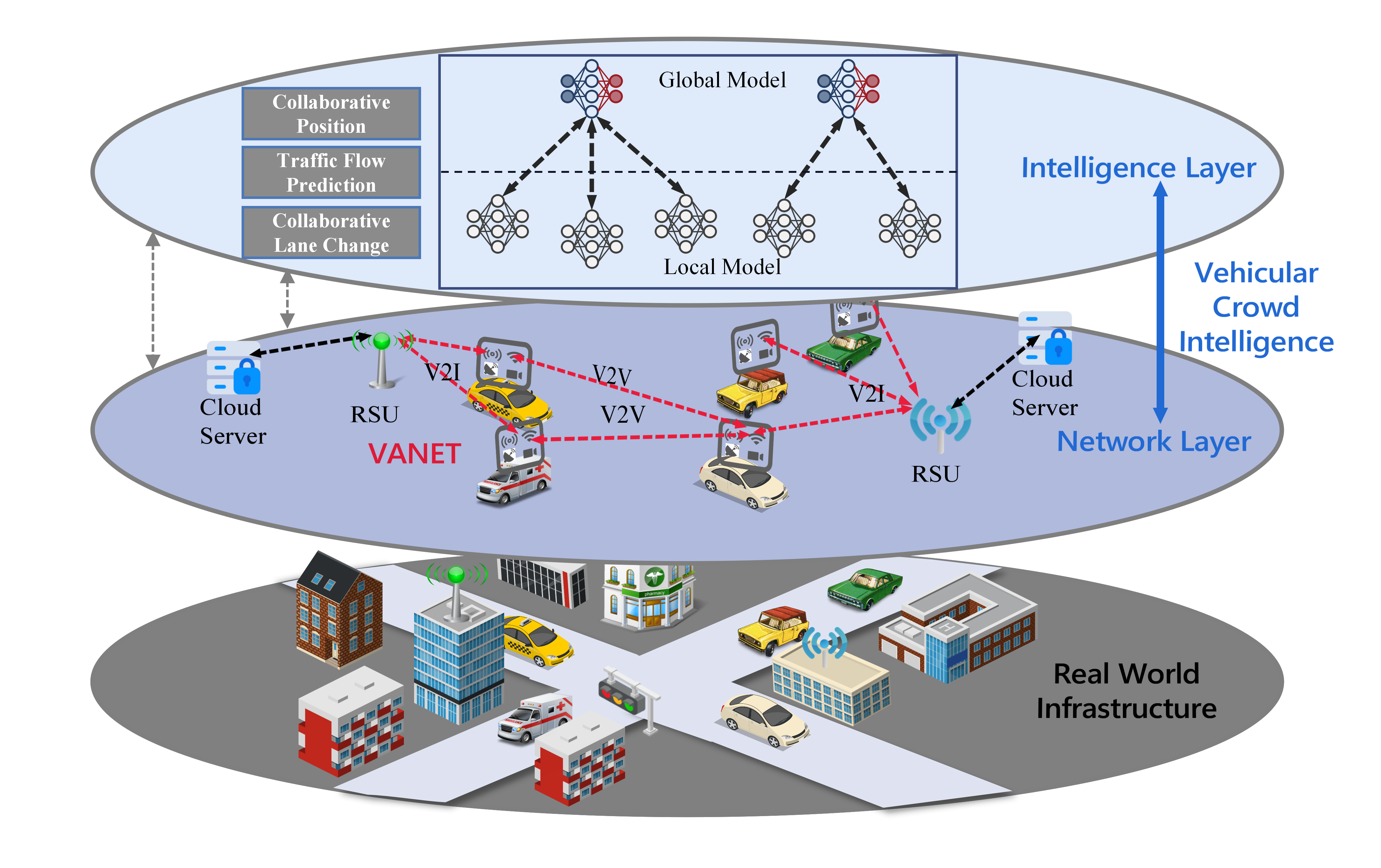}}
\caption{An overview of vehicular crowd intelligence system.}
\label{vci-overview}
\end{figure}

Nowadays, advances in wireless communication technologies, e.g. 5G, contribute to the great success of vehicular ad-hoc networks (VANETs), built upon which such kind of intelligent transportation applications \cite{b1,b11} as traffic accident warning, traffic flow control, road safety analysis, and driving decision making, is delivered to the real world. These applications intrinsically require vehicular nodes to cooperate to produce high-order intelligence, called crowd intelligence.
As shown in Figure~\ref{vci-overview}, vehicular crowd intelligence (VCI) is a two-layer framework that typically comprises multiple intelligent vehicles with various sensors, actuators, and certain computational resources. Enabled by VANETs, these intelligent nodes with AI capacity continuously interact with each other and the surrounding infrastructure to collaboratively find globally, instead of locally, optimal solutions to nontrivial problems.

Despite the promising potential of VCI applications, they suffer from serious data security and privacy issues. On the one hand, intelligent vehicles generate a huge amount of state data and environment-aware data, which usually contain sensitive information, e.g., driving trajectories and pedestrian images. Provided such data are misused or maliciously hacked, there will be a great potential risk of privacy violation. For example, a malicious or even curious-but-honest vehicular node could exploit other nodes' time-series data to obtain private information, such as the location of vehicles and driving behavior\cite{b5}. On the other hand, the collaborative feature of VCI applications yields an urgent need for data sharing; and authorities impose strict security and privacy regulations on data sharing and usage, e.g., GDPR introduced by the European Union, which requires data sharing and usage to be secured.

To tackle this issue, a promising solution is to integrate VCI with the widely-recognized federated learning (FL) paradigm. 
A typical FL framework works as follows: first, a participant is assigned an initial model; second, it updates the model with newly generated data (these data belong to the participant); third, the participant sends local model updates to a remote entity for aggregation and receives back the globally aggregated model.
The approach of using FL is highly dependable as it ensures secure data sharing while maintaining data privacy among participants, which guarantees the formulation of privacy-preserving crowd intelligence that is generally represented by an AI model\cite{msn1}.

However, adapting FL to vehicular crowd intelligence networks remains a challenging problem. As described above, an FL framework involves a remote entity for model aggregation, which is generally instantiated as a trust cloud or edge server. Such a setting is at odds with VCI networks that are highly mobile and delay-sensitive\cite{b12}. More concretely, interacting with a remote server results in an unacceptable amount of time overhead\cite{b13}. As such, due to potential network connectivity issues, either cloud-based or edge-based hybrid solutions cannot deliver a satisfactory privacy-preserving VCI network. Besides, the environment-aware data collected by vehicular nodes are non-iid, i.e., not independent and identically distributed, and it has been shown that non-iid data severely affects the model accuracy in FL\cite{b8}, thus significantly degrading the performance of FL-enhanced VCI networks.

This paper proposes a novel federated learning framework named RaftFed to facilitate privacy-preserving vehicular crowd intelligence. In a nutshell, RaftFed migrates the remote aggregation agent in traditional FL to the VCI network side to address the communication-overhead bottleneck. To this end, it bases a lightweight distributed consensus algorithm to select a vehicular node for model aggregation. Moreover, it opts for a dynamic clustering and hybrid federated learning mechanism to improve Non-IID performance degradation and meanwhile accelerate model convergence.

The contributions of this paper are as follows:
\begin{inparaenum}
\item We propose a lightweight federated learning framework called RaftFed for privacy-preserving vehicular crowd intelligence, which incorporates a two-layer consensus mechanism to effectively select an aggregation node and a dynamic clustering mechanism to mitigate Non-IID performance degradation. 
\item We introduce a pre-training mechanism into RaftFed to exploit non-sensitive data of vehicular nodes so as to improve its convergence rate.
\item We evaluate RaftFed by simulation on well-known public datasets and show that it outperforms baselines with significant performance improvement.
\end{inparaenum}

\section{Related Work }\label{RW}
\noindent\textbf{Vehicular Crowd Intelligence.} 
Collective intelligence, initially introduced by Nehra et al.\cite{b18}, aims to unite a group of individuals with different skills together to handle challenging tasks. Li et.al.\cite{b19} developed this concept by extending the human space with machines, i.e., realizing human-machine collaboration. Later, Yu et al.\cite{crowdsensing} introduced crowd intelligence that deeply integrates human-machine-IoT triple entities to build a self-organizing, self-learning, self-adaptive, and continuous-evolving intelligence space and implement group decision computing in complex scenarios. It is visioned that crowd intelligence could bloom various collaboration-based applications in the AIoT era. This paper targets AIoV (Vehicles), a special field of AIoT, and focuses on investigating how smart vehicles cooperate to handle data-intensive tasks without compromising data privacy. We refer to such a specific context as vehicular crowd intelligence. 

Technically, a VCI system can be described as a network plane atop which an intelligence plane; the vehicular ad-hoc network technique is a cornerstone of the VCI network plane, which enables vehicular nodes to communicate with each other and the surrounding environment\cite{b21}. In the literature, VANETs have been well-studied regarding data routing, message broadcasting, and quality of communication service, but data security and privacy issues are still not well-solved since trade-offs between performance and security are nontrivial\cite{vanet-sec}.

\noindent\textbf{VCI-oriented Federated Learning Schemes.}
Google first introduced Federated learning in 2016; it aims to train a globally shared model among distributed mobile devices for image classification and language word prediction\cite{b9}. FL's typical workflow is as follows: the remote server provides an initial model and distributes it to all clients for local training; after the local training process on the client side is completed, the local model's parameters will be sent to the remote server for aggregation, thus obtaining a global model. Training models locally could significantly eliminate the privacy risks of raw data and greatly reduce communication costs among involved entities. As such, FL has been widely applied in various collaboration-based applications that require data privacy protection \cite {b14}, especially for VCI.

Samarakoon et al. \cite{fedurll} first proposed an FL-based resource allocation method for VANETs to deliver ultra-reliable, low-latency communications. Specifically, their work exploits FL to reduce the communication overhead caused by sharing statistics required by the URLL mechanism while preserving data privacy. Li et al. \cite{fedvanet} introduced a clustered FL scheme to the general applications in VANET, which embeds an intra-cluster recursive training mechanism and an inter-cluster cycling update mechanism. Afterward, they based the prior scheme to additionally propose a Paillier algorithm to enhance the security of parameters transmitted among involved entities\cite{b8}. Kong et al.\cite{fedvcp} implemented privacy-preserving cooperative positioning by creating an FL-enabled VCI network where sensor-rich vehicles are utilized to securely share the positioning information with other common ones. 

However, all existing work opts for a remote entity-based FL framework; communications with the remote entity, either a cloud or an edge server, yield intolerable overheads and unreliable channels in view of VCI applications that are highly mobile and time-delay-sensitive. How to effectively adapt FL in VCI networks remains an open problem.

\section{Problem Formulation}\label{PF}
\begin{figure}[!ht]
	\centerline{\includegraphics[scale=0.06]{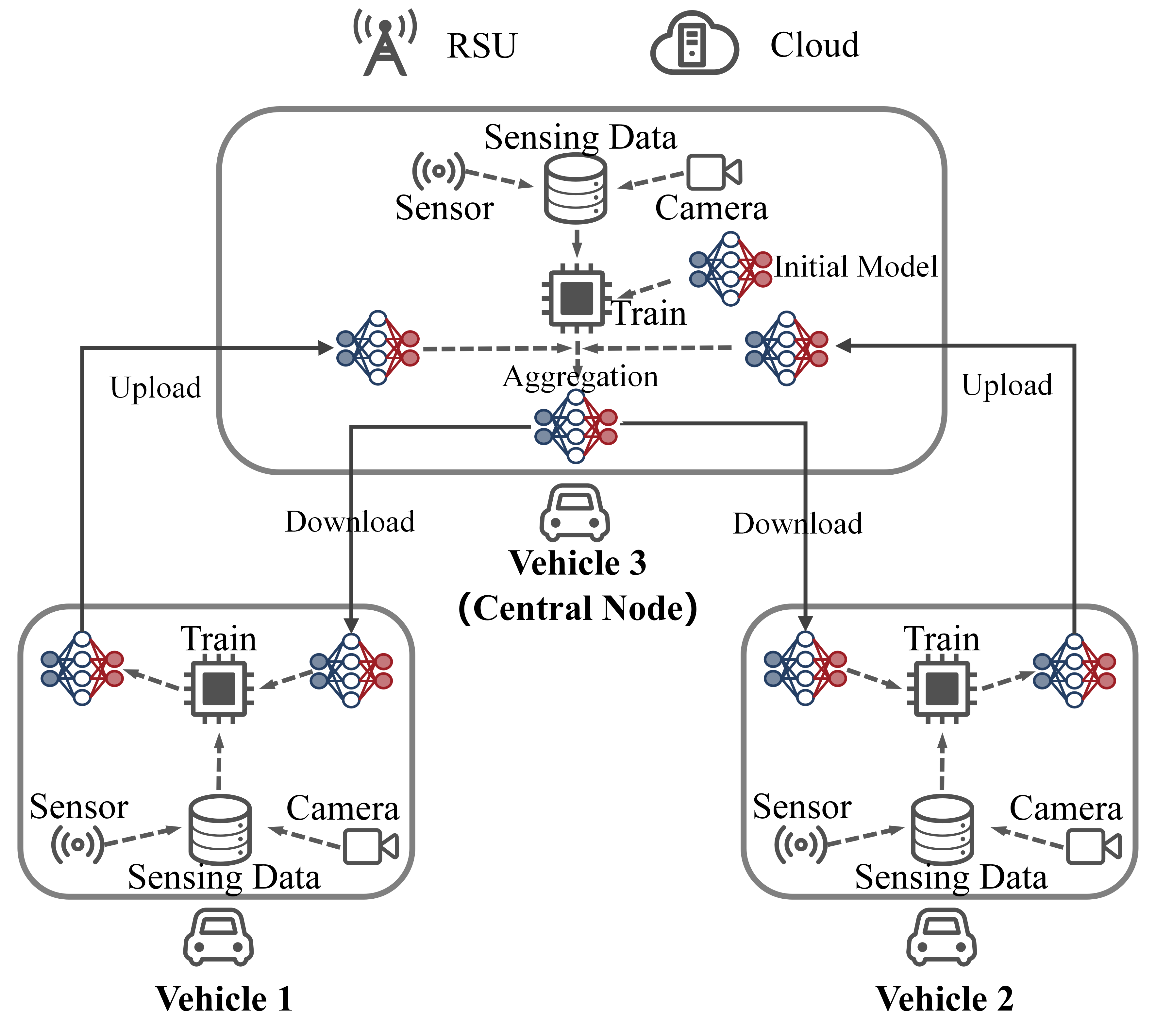}}
	\caption{A lightweight VCI-oriented federated learning framework.}
	\label{vci-fl-model}
\end{figure}

A conventional VCI-oriented federated learning framework consists of a remote server (RSU or cloud) and several clients (vehicular nodes). In this paper, by contrast, we aim to propose a lightweight federated learning scheme for VCI. As shown in Figure~\ref{vci-fl-model}, such a framework only involves vehicular nodes; in other words, both model training and aggregation are finished by the vehicular network itself without communicating to a remote RSU or the cloud. More specifically, each vehicular node in the network trains a model locally, then sends the updated parameters to one (including the sender itself) of the network nodes for aggregation, and then receives back the parameters of the aggregated model for the next iteration. The local training process and aggregation process are respectively modeled as:

$$w_{client}^{t+1}=\boldsymbol{LocalTrain}(w_{agg}^{t},D_{client})$$

$$w_{agg}^{t+1}=\boldsymbol{Aggregate}(\bigcup_{i=1}^{|client|}w_{client}^{t+1})$$

\noindent where $t \in [0,n]$ represents iteration round, $w_{agg}^{t}$ denotes the global model parameters of $t$-th training round, $w_{client}^{t}$ denotes the local model parameters of $t$-th training round, and $D_{client}$ denotes the newly private data of the current vehicular node. After $n$ epochs, a converged global model will be obtained. 

Without loss of generality, we formally define an FL-based VCI task as an optimization problem that aims to minimize the empirical loss of a neural network model trained by multiple vehicular nodes in a distributed manner. 

Given an FL-based VCI task, let $V$ denote a set of vehicular nodes in the VCI network, where each node $v_i$ has a certain amount of sensing data $x_i$ along with labels $y_i$; $i \in [1 \cdots |V|]$ and $|V|$ denotes the number of involved vehicular nodes. Typically, each node is equipped with a TEE-supported CPU and a wireless network device to securely perform specific intra-node or inter-node tasks. 

The target optimization problem is formulated as:
$${\rm argmin}\ F(w)=\sum_{i=1}^{|V|}p_{i}f(w,D_i^v)$$
where $p_{i}$ denotes weights used in aggregation, $F(w)$ denotes the empirical loss of a model represented by unique parameters $w$, $ D_i^v$ denotes the training data of the $i$-th node and $D_i^v$ = \{($x_i$, $y_i$) $|$ $i \in [1, \dots, |V|]$\}, the $argmin$ function is to find a set of parameters that makes the empirical loss of global model minimum, and $f(\cdot)$ is an empirical loss function for a local model trained by a vehicular node, which is defined as:

$$f(w,D_i^v)=\frac{1}{|D_i^v|}\sum_{j=1}^{|D_i^v|}\mathcal{L}(w,D_{i,j}^v)$$

$$s.t.\ D_i^v=\bigcup_{j=1}^{|D_{i}^v|}D_{i,j}^v$$
where $\mathcal{L}(w,D_{i,j}^v)$ denotes a loss function of $j$-th batch of $D_{i}^v$, e.g., cross-loss entropy.

\begin{algorithm}[!h]
	\caption{\textbf{VCI-oriented Efficient Federated Learning}}
        \label{algo-raftfed}
	\begin{algorithmic}[1]
            \REQUIRE Learning Rate $\eta$ , Local Training Round $E$ , Global Training Round $T$ , Pre-Training Round $T_p$ ,Inter-Cluster Training Round $T_i$
		\STATE \textbf{Dynamic Clustering Phase:}
		\STATE  $L_{(cl)} \leftarrow $Vehicular Set $L_{(v)}$ executes \textbf{Algorithm~\ref{algo-cluster}} 
		\FOR{$cluster$ in $L_{(cl)}$}		
            \FOR{vehicle $v_{(i)}$ in $cluster$}		
		\IF{$v_{(i)}$ is non-private vehicle}		
		\STATE $v_{(L)}^i \Leftarrow$  non-private data $\widetilde D_k^i$ of $v_{(i)}$		
				\ENDIF
			\ENDFOR
		\ENDFOR
		\STATE $v_{(LF)} \leftarrow $ Intra-cluster aggregation nodes set $L_{(c)}$ executes \textbf{Algorithm~\ref{algo-inter}}  		
		\FOR{$cluster$ in $L_{(cl)}$}		
            \STATE $v_{(LF)} \Leftarrow$  non private data $\widetilde D_{k}$  of $cluster$	
            \STATE  $v_{(LF)} \Leftarrow$ number of training data samples $\widehat D_{k}$  of 	$cluster$	
		\ENDFOR
		\STATE \textbf{Model Training Phase:}
		\FOR{$t = 1,\dots,T_p$ }		
            \STATE $v_{(LF)}$ pre-training initial global model $M$ 		
		\ENDFOR
		\STATE $v_{(LF)}$ send initial model parameter $\hat \theta _0 $ and training sequence	to first $cluster$ 	
		\FOR {$t = 1,\dots,T$ }
            \FOR{$cluster$ in $L_{(cl)}$}		
            \STATE $\theta _{t-1} \leftarrow  \hat \theta _{t-1} $		
            \FOR{$epoch = 1,\dots,T_i$}
            \FOR{vehicle $v_{(i)}$ in $cluster$}		
            \STATE  receive model and training sequence from $v_{(i-1)}$ 	
            \FOR{$epoch = 1,\dots,E$}											
            \STATE  $v_{(i)}$ training local model $M^i$ 		
            \STATE  $\theta _t^i  \leftarrow  \theta _{t-1}^i - \eta \nabla F^i$ 		
            \ENDFOR
            \STATE $\theta _t^i  \leftarrow  \frac{|D_{k} |} {|  D |}\theta _{t}^i +\left ( 1-\frac{| D_{k} |} {|  D |}\right )\theta _{t-1}^i$		
            \STATE $v_{(j)}\Leftarrow$$\theta _t^i$ 		
				\ENDFOR
				\ENDFOR
			\STATE $\theta _t \leftarrow  \theta _t^i$ 		
			\STATE next cluster $\Leftarrow \theta _t$ 		
			\STATE $\hat \theta _{t}  \leftarrow   \frac{|\widehat D_{k} |} {| \widehat D |}\theta _{t} +\left ( 1-\frac{|\widehat D_{k} |} {| \widehat D |}\right ) \theta _{t-1}$ 
			\ENDFOR
		\ENDFOR
	\end{algorithmic}
\end{algorithm}

\section{The Framework of RaftFed}\label{RaftFed}
In this section, we present the details of RaftFed, the workflow of which is illustrated in Algorithm~\ref{algo-raftfed}. 

Recall that the core idea of RaftFed is to make a participating node instead of a remote server handle model aggregations, so RaftFed must answer how to efficiently identify the specific node. Besides, given the Non-IID vulnerabilities of FL, RaftFed must also answer how to make such a node identification effective. RaftFed bases a distributed consensus algorithm for the first question to negotiate an aggregator among vehicular nodes. For the second question, it categorizes vehicular nodes according to their data distribution by exploiting the VANET clustering mechanism; correspondingly, it extends the single-negotiation process to a dual-negotiation one so as to deliver an effective intra-cluster model aggregation and efficient inter-cluster model aggregation. As such, RaftFed comprises two phases: cluster election that forms a two-layer aggregation structure and model training that accomplishes federated learning procedures. Note that the aggregation processes are assumed to be securely performed and this can be delivered by seamlessly integrating out-of-box hardware enclaves.

\begin{algorithm}[!h]
	\caption{\textbf{Intra-cluster Aggregation Node Election}}
        \label{algo-leader}
	\begin{algorithmic}[1]
            \REQUIRE Vehicular Set $L_{(v)}$
            \ENSURE Intra-cluster Aggregation Node $v_{(L)}$
            \FOR {each vehicle $v_{(i)}$}		
            \IF{$v_{(i)}$did not choose to be a candidate}		
            \STATE $v_{(i)}$ become Follower $F_i$			
            \ELSE
            \STATE $v_{(i)}$ send $Heartbeat$ to all other vehicle $v_{(j)}$		
            \ENDIF
        
            \ENDFOR
            \STATE $Follower$ receives all $Heartbeat$ from $Candidate$	
            \FOR{Follower $F_i$ in vehicle set $L_{(v)}$}		
            \STATE Votes for the first candidate who sends $Heartbeat$		
            \ENDFOR
            \FOR{Candidate $C_i$ in vehicle set $L_{(v)}$}		
            \IF{$count_{C_i} > 1/2 |L_{(v)}|$}			
            \STATE $C_i$ become the Intra-cluster Aggregation Node $v_{(L)}$			
            \STATE $v_{(L)}$ send message to all other vehicles $v_{(i)}$ 		        
            \ENDIF
            \IF{$count_{C_i} <= 1/2 |L_{(v)}|( i = 1, \dots , |L_{(v)}|$)}
            \STATE return step 1			
            \ENDIF
            \ENDFOR
            \STATE  Return $v_{(L)}$ 
	\end{algorithmic}
\end{algorithm}

\subsection{Clustering Election Phase}
A distributed consensus algorithm and a dynamic clustering algorithm are respectively developed in the clustering election phase. The distributed consensus algorithm aims to negotiate a leader among a group of vehicular candidates for clustering assistance or model aggregation; the dynamic clustering algorithm aims to categorize vehicular nodes into different groups based on their data distribution. Built upon these two algorithms, the clustering election phase works through the following three steps:
\begin{inparaenum}
    \item Identifying a leader node from the initial vehicular set with the distributed consensus algorithm;
    \item The leader node assists in clustering vehicular nodes with the dynamic clustering algorithm and meanwhile yielding intra-cluster aggregation nodes;
    \item Electing a global (i.e., inter-cluster) aggregation node from intra-cluster aggregation nodes.
\end{inparaenum}

Thus, a two-layer aggregation structure is established among participating vehicular nodes in a VCI network. Next, we detailedly introduce the two algorithmic building blocks.

\subsubsection{The Distributed Consensus Algorithm}

\begin{figure}[!h]
	\centerline{\includegraphics[scale=0.045]{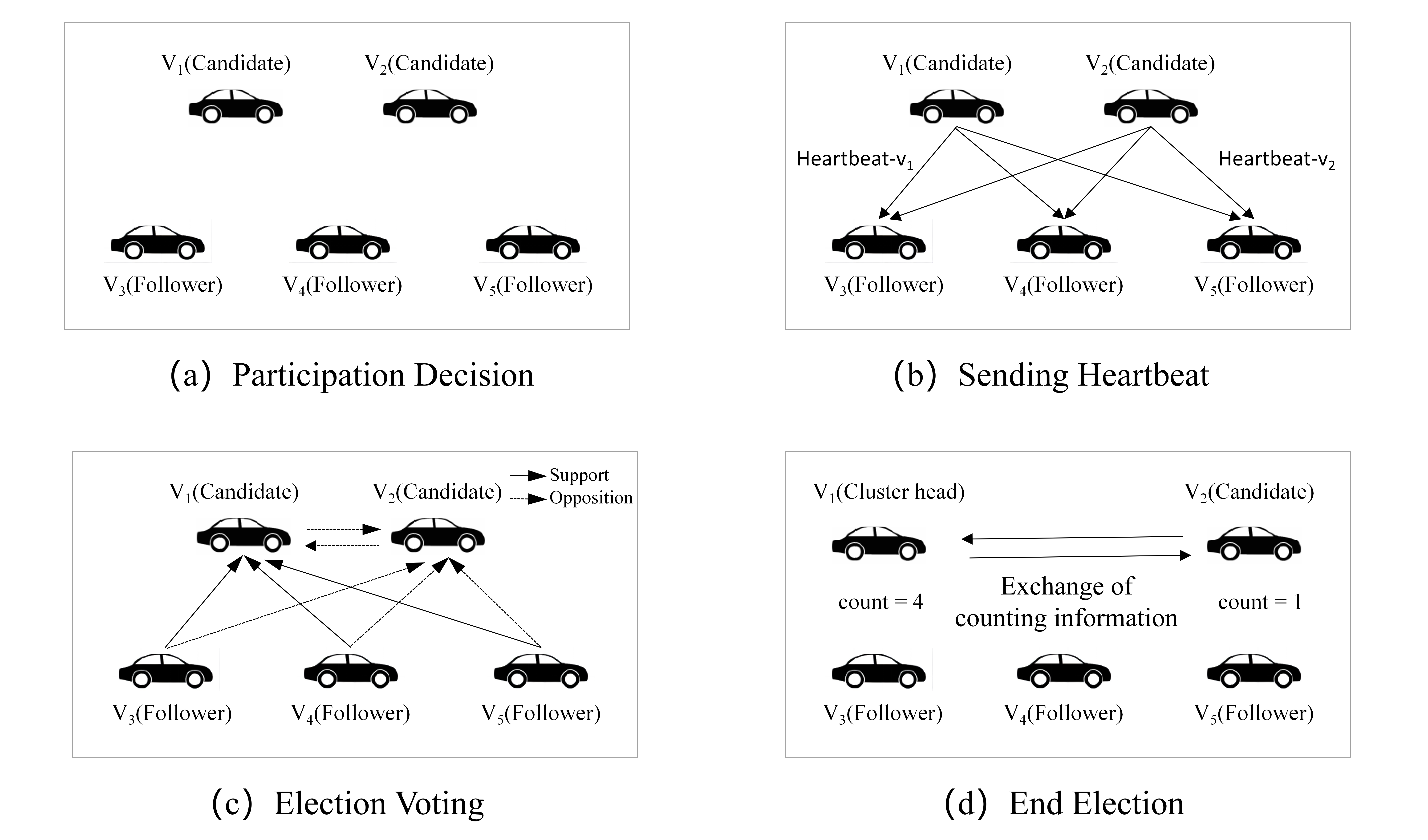}}
	\caption{The distributed consensus procedure.}
	\label{fig-consensus}
\end{figure}

As illustrated in Algorithm~\ref{algo-leader} and Algorithm~\ref{algo-inter}, the distributed consensus algorithms accomplish the selection of the leader node and the global aggregation node, respectively. Please note that we divide the role of vehicles into Follower, Candidate, and Leader to realize the election of vehicle nodes, where the Leader is a node to be elected out, Candidate is a participant, and Follower is a node that responds to candidates and makes a vote.

The consensus process behind the two algorithms mainly includes the following steps, shown in Figure~\ref{fig-consensus}. 

\begin{algorithm}[!h]
	\caption{\textbf{Global Aggregation Node Election}}
        \label{algo-inter}
	\begin{algorithmic}[1]
            \REQUIRE Intra-cluster aggregation nodes set $L_{(c)}$, non-sensitive data $\widetilde D_k^i$
            \ENSURE Global aggregation node $v_{(LF)}$
            \FOR{each intra-cluster aggregation nodes $v_L^i$ from 1 to $n$}		
            \IF {$v_L^i$ did not choose to be a candidate $C_i$}			
            \STATE $v_L^i$ become Follower $F_i$			
            \ELSE
            \STATE $v_L^i$ send $Heartbeat$ and   $|\widetilde D_k^i|$ to all other intra-cluster aggregation nodes $v_L^j$ 		
            \ENDIF
            \ENDFOR
            \STATE $Follower$ receives all $Heartbeat$ from $Candidate$		
            \FOR{Follower $F_i$ in Intra-cluster aggregation nodes set $L_{(c)}$}			
            \STATE Votes for the $C_i$ with the largest $|\widetilde D_k^i|$			
            \ENDFOR
            \FOR{Candidate $C_i$ in Intra-cluster aggregation nodes set $L_{(c)}$}				
            \IF{$count_{C_i} > 1/2 |L_{(c)}| $}				
            \STATE $C_i$ become the Global aggregation node $v_{(LF)}$			
            \STATE $v_{(LF)}$ send message to all other intra-cluster aggregation nodes $v_L^i$  			
            \ENDIF
            \ENDFOR
            \IF{$count_{C_i} <= 1/2 |L_{(c)}|( i=1 ,\dots ,|L_{(c)}|$) }			
            \STATE return step 1				
            \ENDIF
		\STATE Return $v_{(LF)}$; %算法的返回值
	\end{algorithmic}
\end{algorithm}

\paragraph{\textbf{Participation Decision}} 
The vehicle $v_i$ in the vehicle set $L_{(v)}$ chooses whether to participate or not. If yes, the role of the $v_i$ becomes $\textbf{Candidate}$, otherwise $\textbf{Follower}$. Candidate $C_i$ has \textbf{one} vote and votes for itself by default(see Algorithm~\ref{algo-leader} line 1-3, Algorithm~\ref{algo-inter} line 1-3).

\paragraph{\textbf{Heartbeat Sending}} 
In the leader node election, all candidates will send a $Heartbeat$ signal to others (see Algorithm~\ref{algo-leader}, line 5), while in the inter-cluster aggregation node election, all candidates send $Heartbeat$ signal and the amount of non-sensitive data about their clusters (see Algorithm~\ref{algo-inter}, line 5), such that all vehicles know the identity of the candidate.

\paragraph{\textbf{Election Voting}} 
Every follower $F_i$ elects an appropriate candidate according to the corresponding election principles. In the leader node election, every follower $F_i$ votes for the candidate that first sends the message (see Algorithm~\ref{algo-leader}, line 8-11), while in the inter-cluster aggregation election, follower $F_i$ will preferentially vote for the Candidate with the largest amount of non-sensitive data $\widetilde D_k^i$ (Algorithm~\ref{algo-inter} line 8-11). $F_i$  will send a $pro-vote$ message to the corresponding candidate and an $anti-vote$ message to other candidates.

\paragraph{\textbf{End Election}} 
All candidates count their votes and exchange counting information with each other. The election rules are set as follows:

$$count >1/2|L_{(v)}|$$

If a candidate $C_i$ satisfies the election rules, $C_i$ will become the elected node $v_{(L)}$, and the algorithm ends; otherwise, the consensus algorithm is re-executed (see Algorithm~\ref{algo-leader} line 12-18, Algorithm~\ref{algo-inter} line 12-19).

\begin{algorithm}[!h]
	\caption{\textbf{ Dynamic Clustering}}%算法标题
	\label{algo-cluster}
	\begin{algorithmic}[1]%一行一个标行号
		\REQUIRE Vehicular Set $L_{(v)}$, Data Distribution Overlap Rate Lower Limit $R$ 
		\ENSURE ClusterSet $L_{(cl)}$
		\STATE Initial cluster $C_{(hi)} \leftarrow L_{(v)}$ 		
		\STATE $C_{(hi)}$ joins ClusterSet $L_{(cl)}$ 		
		\STATE  $v_{(L)} \leftarrow$$L_{(cl)}$ executes \textbf{Algorithm~\ref{algo-leader}} 		
		\FOR{vehicle $v_{(L)}$ in $C_{(hi)}$}		
		\FOR{$cluster$ in $L_{(cl)}$}			
		\STATE Calculate the $w$ of the $v_{(L)}$ and $v_{(L)}^i$ 	
		\IF{$w \geq R$}		
		\STATE $v_{(L)}$ joins $cluster$ 		
		\ENDIF
		\ENDFOR
		\IF{$v_{(L)}$ can't join any  $cluster$}		
		\STATE create $newCluster$ and $v_{(L)}$ to be the intra-cluster aggregation nodes $v_{(L)}^j$ 		
		\STATE $newCluster$ joins $L_{(cl)}$ 		
		\ENDIF
		\ENDFOR
		\STATE Return $L_{(cl)}$
	\end{algorithmic}
\end{algorithm}

\subsubsection{The Dynamic Clustering Algorithm}
As illustrated in Algorithm~\ref{algo-cluster}, the dynamic clustering algorithm divides the initial vehicular set $V$ into multiple clusters and chooses all intra-cluster aggregation nodes, which mainly includes the following steps.

\paragraph{\textbf{Leader Node Election}} 
RaftFed initializes an initial cluster set $C_{(hi)}$ with all vehicular nodes, whose cluster head $v_{(L)}$, i.e., the leader node, is elected by Algorithm~\ref{algo-leader}. This node is then responsible for instructing the generation of new clusters $L_{(cl)}$ (see Algorithm~\ref{algo-cluster} line 2-3).

\paragraph{\textbf{Dynamic Clustering}} 
In this step, it analyzes the distribution of training data for all remaining vehicular nodes in $C_{(hi)}$. To this end, we use the data distribution overlap rate to measure the distribution difference between the leader node and $v_{(L)}^i$, which is formulated as
$$R=\frac{D_{v_i}^y}{D_{v_{(L)}^i}^y}$$
where $D_{i}^y$ represents the number of data labels of node $i$. We define the lower bound of the data distribution overlap rate of clusters. A vehicular node $v_{(L)}^j$ traverses the current set of clusters. When the data distribution overlap rate between the vehicular node and a cluster aggregation node is greater than the threshold, node $v_{(L)}^j$ will be added to the cluster. If the above conditions are not met after traversing all clusters, node $v_{(L)}^j$ creates a new cluster $newCluster$ and is treated as the cluster head, i.e., intra-cluster aggregation node, and cluster $newCluster$ is added to the set $L_{(cl)}$ (Algorithm~\ref{algo-cluster} line 4-13).

\paragraph{\textbf{Inter-cluster Node Election}} 
After dynamic clustering, the global (inter-cluster) aggregation node needs to be elected to perform global model aggregation. More specifically, a small portion of data from non-private nodes $D_k^i$ are first collected by each cluster's aggregation node. Then, an inter-cluster node $v_{(LF)}$ is elected among intra-cluster aggregation nodes according to the amount of such non-sensitive data by Algorithm~\ref{algo-inter} (Algorithm~\ref{algo-inter} lines 8-19). Please note that we refer to vehicular nodes that do not contain any sensitive information in a cluster as non-private nodes, e.g., buses.

\subsection{Model Training Phase}
When the clustering election phase is finished, a two-layer aggregation structure for VCI networks is set up. And then, the model training phase bases such a setting to collaboratively build network-wide knowledge among vehicular nodes by exchanging model updates, which specifically includes the following steps. 

\subsubsection{\textbf{Model Pre-training}} 
RaftFed integrates a pre-training mechanism to improve the model convergence rate by (see Algorithm~\ref{algo-raftfed} lines 11-18): 
\begin{inparaenum}
    \item aggregating non-sensitive data at the global aggregation node side to initialize a pre-training auxiliary dataset; and
    \item with which to produce an initial global model by the global aggregation node.
\end{inparaenum}

\subsubsection{\textbf{Model Training}} 
Typically, the model training step includes two stages i.e., iteration-based intra-cluster training and loop-based inter-cluster training. 
RaftFed opts for a loop-based training approach instead of a traditional iteration-based one among clusters to further reduce the communication overhead. More specifically, the global aggregation node first sends the initial model to one of the intra-cluster aggregation nodes and then specifies the order of clusters for loop training. 

\paragraph{\textbf{Intra-Cluster Training}}
Within each cluster, the aggregation node coordinates model training by the well-known Stochastic Gradient Descent (SGD) algorithm, denoted by
$$\theta_t^i=\theta_{t-1}^i-\eta\nabla F^i$$
where $\theta_t^i$ represents model parameters in the $t$-th epoch for $i$-th vehicular node and $\eta$ represents the learning rate. In RaftFed, the model parameters will be aggregated for each vehicular node within a cluster after $E$ training epochs, which is computed by
$$\theta_t^i=\frac{|D_k|}{|D|}\theta_t^i+(1-\frac{|D_k|}{|D|})\theta_{t-1}^i$$
where $|D_k|$ represents the number of training data samples of $k$-th vehicular node, and $|D|$ represents the number of training data samples of the cluster.

\paragraph{\textbf{Inter-Cluster Training}}
Given a vehicular cluster, it will receive a model update from the intra-cluster aggregation node of its predecessor cluster and aggregate the update using
$$\theta_t^i=\frac{|{\hat{D}}_k|}{|D|}\theta_t^i+(1-\frac{|{\hat{D}}_k|}{|D|})\theta_{t-1}^i$$
where $|{\hat{D}}_k|$ represents the number of training data samples of the $k$-th cluster.

\section{Performance Evaluation}\label{PE}

\begin{figure}[!h]
	\centerline{\includegraphics[scale=0.34]{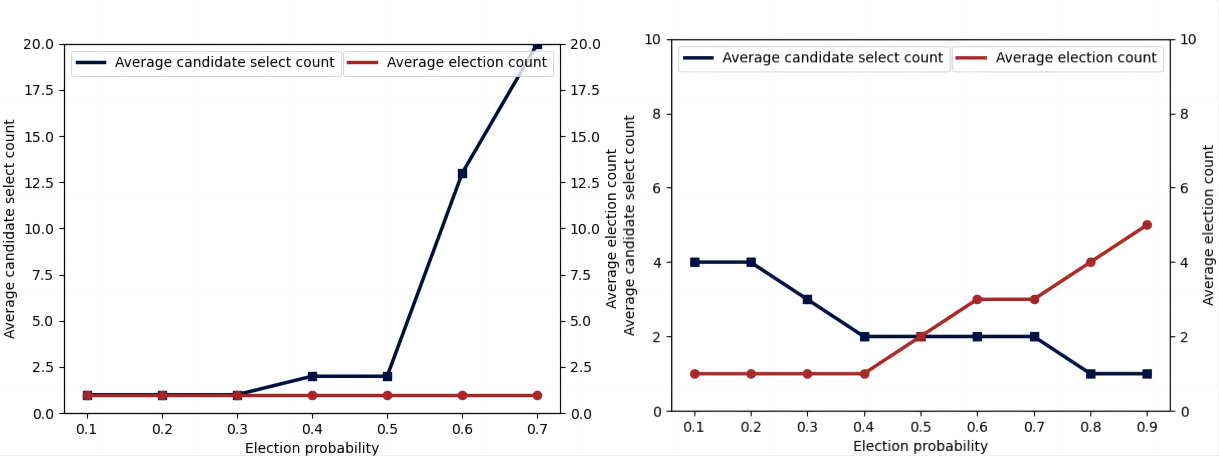}}
	\caption{Comparison of election rounds with different joining probabilities.}
	\label{fig-base-robustness}
\end{figure}

\subsection{Optimal Hyperparameters for Election Voting} 
The election voting algorithm involves a node participating in an election with a specific probability. To determine the optimal value, a pretest is conducted beforehand.
We tested with 10 and 100 nodes and examined the number of rounds required to choose candidates and the success rate of elections at varying joining probabilities.

It can be seen from Figure~\ref{fig-base-robustness} that when the joining probability of each node is low, it generally needs to launch several elections to find a central node; when the joining probability is more than 0.6, the number of candidates will exceed the half of the total number of nodes, and in this case, it is impossible to successfully finish election since the criterion of electing a center node cannot be satisfied. We, therefore, propose to set the joining probability of nodes as 0.5.

\subsection{RaftFed's Performance}
\noindent\textbf{Settings:} A simple CNN model for image classification with the MNIST dataset and a ResNet18\cite{resnet18} one with the CIFAR-10 dataset are used as experimental tasks.
The naive model has two convolutional layers: the first layer contains 1 input channel and 32 output channels with a $5 \times 5$ convolutional kernel size, step-size = 1, and padding = 2; the second layer contains 32 input channels and 64 output channels with a $5 \times 5$ convolutional kernel size, step-size = 1, and padding = 2. Each convolutional layer is followed by a $2 \times 2$ maximum pooling layer; the two layers are connected with out-feature = 512 and 10, respectively.
As shown in Table~\ref{dataset-setting}, we manually divide the MNIST data into Non-iid distribution patterns.
As for CIFAR-10, we use random allocation to distribute the data and ensure that each node does not contain all the labeled data and meanwhile, we set the percentage of shared data as 5\%.
Besides, we set the pre-training round as 50, the data sharing ratio as 5$\%$, and the batch size as 20 in the pre-training setup, using the stochastic gradient descent (SGD) method with the learning rate = $10^{-5}$, and set the number of global training rounds to 100 and the number of training rounds to 2 within the cluster in the formal training. 

\noindent\textbf{Baselines:} For MNIST,we used state-of-the-art FedCluster\cite{fed-cluster}, FedVANET\cite{fedvanet}, and Semi-FL\cite{semi} as the baselines; For CIFAR-10, we opted in FedVANET as the baseline.

\begin{table}
\centering
\caption{The distribution of MNIST Dataset with Non-IID setting .}
\label{dataset-setting}
\scalebox{1.13}{
\begin{tblr}{
  row{even} = {c},
  row{3} = {c},
  row{5} = {c},
  cell{1}{1} = {c},
  cell{1}{2} = {c},
  cell{1}{3} = {c},
  hline{1,7} = {-}{0.08em},
  hline{2} = {-}{},
}
Client & Traning-Data Labels & Client & Traning-Data Labels \\
0      & \textbf{0,4,6,8,9}  & 5      & \textbf{3}          \\
1      & \textbf{1,4,5,6}    & 6      & \textbf{2,3,5}      \\
2      & \textbf{0,6,7,9}    & 7      & \textbf{3,5}        \\
3      & \textbf{1,2,0,9}    & 8      & \textbf{7,8}        \\
4      & \textbf{1,2,4,8}    & 9      & \textbf{7}          
\end{tblr}
}
\end{table}

\begin{table}
\centering
\caption{Comparison of accuracy between RaftFed and baselines on MNIST}
\label{accuracy-res-mnist}
\scalebox{1.18}{
\begin{tblr}{
  cells = {c},
  cell{1}{1} = {r=2}{},
  cell{1}{2} = {r=2}{},
  cell{1}{3} = {c=3}{},
  hline{1,7} = {-}{0.08em},
  hline{3} = {-}{},
}
Rounds & RaftFed          & Baseline   &         &          \\
      &                  & FedCluster & Semi-FL & FedVANET \\
25    & \textbf{96.29\%} & 79.34\%    & 85.22\% & 93.71\%  \\
50    & \textbf{97.66\%} & 90.80\%    & 86.78\% & 96.01\%  \\
75    & \textbf{98.00\%} & 92.19\%    & 87.56\% & 97.33\%  \\
100   & \textbf{98.30\%} & 93.86\%    & 88.06\% & 97.54\%  
\end{tblr}
}
\end{table}

% \begin{table}
% \centering
% \caption{Comparison of loss between RaftFed and baselines}
% \scalebox{1.23}{
% \begin{tblr}{
%   cells = {c},
%   cell{1}{1} = {r=2}{},
%   cell{1}{2} = {r=2}{},
%   cell{1}{3} = {c=3}{},
%   hline{1,7} = {-}{0.08em},
%   hline{3} = {-}{},
% }
% Round & RaftFed         & Baseline   &         &          \\
%       &                 & FedCluster & Semi-FL & FedVANET \\
% 25    & \textbf{0.7741} & 3.0141     & 2.7955  & 1.1568   \\
% 50    & \textbf{0.5197} & 1.9289     & 2.6791  & 0.7837   \\
% 75    & \textbf{0.4468} & 1.4434     & 2.5147  & 0.6041   \\
% 100   & \textbf{0.3958} & 1.1072     & 2.3777  & 0.4943   
% \end{tblr}
% }
% \end{table}

\begin{figure}[!h]
\centerline{\includegraphics[scale=0.23]{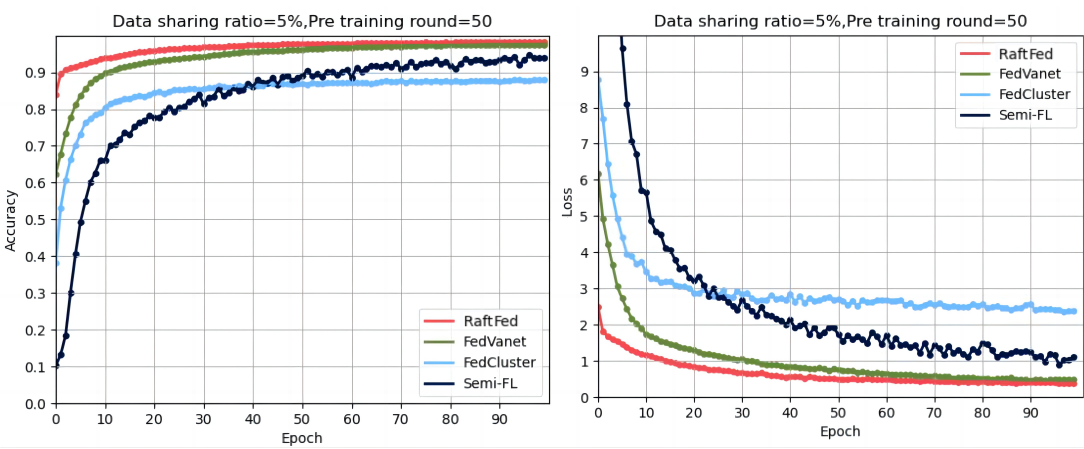}}
\caption{Comparison of accuracy and loss between RaftFed and baselines on MNIST.}
\label{fig-base-accuracy-mnist}
\end{figure}

\begin{figure}[!h]
	\centerline{\includegraphics[scale=0.23]{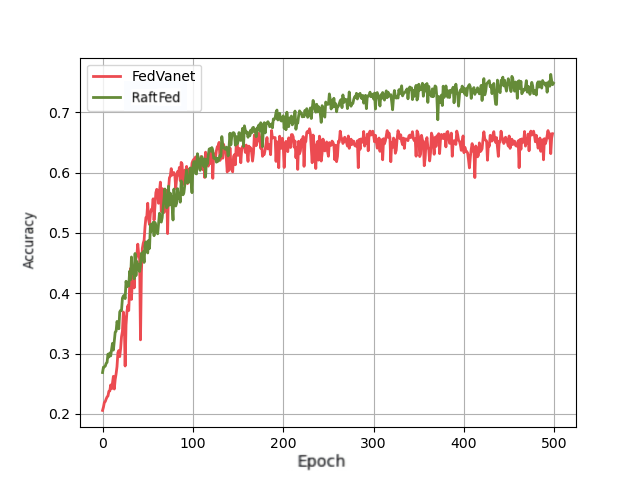}}
	\caption{Comparison of accuracy between RaftFed and baselines on CIFAR-10.}
	\label{fig-base-accuracy-cifar10}
\end{figure}

\begin{figure}[!h]
\centerline{\includegraphics[scale=0.23]{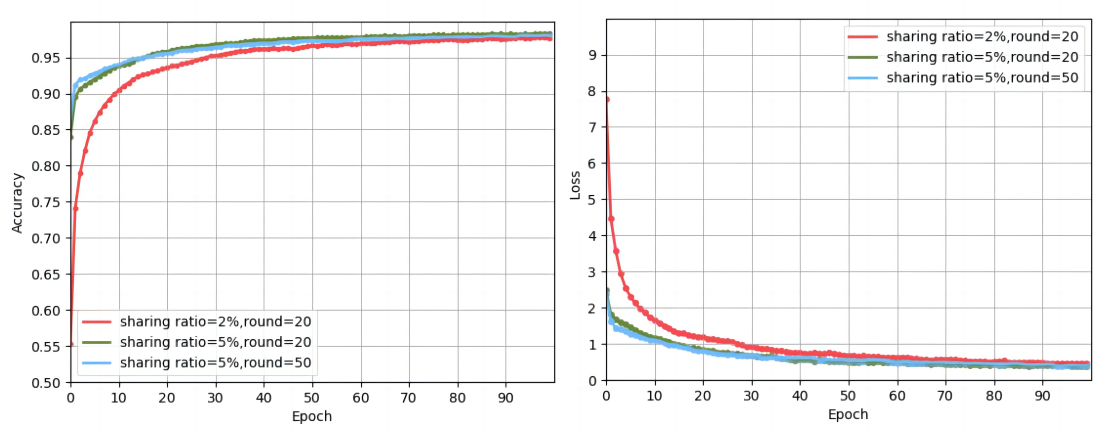}}
\caption{RaftFed performance under different pre-training rounds and data sharing ratio conditions with MNIST.}
\label{fig-self}
\end{figure}

\noindent\textbf{Results:} The accuracy comparison of our framework with other baselines on the MNIST dataset is shown in Table~\ref{accuracy-res-mnist} and Figure~\ref{fig-base-accuracy-mnist}. Our method has 96.29$\%$ of accuracy at epoch = 25, followed by FedVanet with 93.71$\%$, Semi-FL with 85.22$\%$, and FedCluster with only 79.34$\%$; It has 98.30$\%$ of accuracy at epoch = 100, followed by FedVanet with 97.54$\%$, FedCluster with only 93.76$\%$, and Semi-FL with 88.06$\%$. 
Regarding convergence speed, RaftFed outperforms FedVanet and substantially precedes Semi-FL and FedCluster. 
Pertaining to the Cross-Entropy loss of the task, the results are consistent with the above findings, and at epoch = 100, RaftFed has the lowest loss. Table~\ref{accuracy-res-cifar10} and Figure~\ref{fig-base-accuracy-cifar10} demonstrate the experimental results with the CIFAR-10  dataset. RaftFed has an accuracy of 75.83\% when round = 500, which is around 9\% better than the baseline.
Figure~\ref{fig-self} shows how RaftFed performs on the MNIST dataset with different experimental settings. We conclude that a small portion of data sharing and model pre-training will significantly improve the performance and convergence efficiency of the model.

\begin{table}
	\centering
	\caption{Comparison of accuracy between RaftFed and Baselines on CIFAR-10}
	\label{accuracy-res-cifar10}
	\begin{tblr}{
			column{2} = {c},
			column{5} = {c},
			column{8} = {c},
			cell{1}{1} = {r=2}{},
			cell{1}{2} = {r=2}{},
			cell{1}{3} = {r=2}{},
			cell{1}{4} = {r=2}{},
			cell{1}{5} = {r=2}{},
			cell{1}{6} = {r=2}{},
			cell{1}{7} = {r=2}{},
			hline{1,8} = {-}{0.08em},
			hline{3} = {-}{},
		}
		& Rounds &  &  & RaftFed          &  &  & Baseline         &  \\
		&        &  &  &                  &  &  & FedVANET         &  \\
		& 100    &  &  & 61.86\%          &  &  & \textbf{62.15\%} &  \\
		& 200    &  &  & \textbf{69.31\%} &  &  & 64.85\%          &  \\
		& 300    &  &  & \textbf{73.23\%} &  &  & 65.12\%          &  \\
		& 400    &  &  & \textbf{74.29\%} &  &  & 64.49\%          &  \\
		& 500    &  &  & \textbf{75.83\%} &  &  & 65.61\%          &  
	\end{tblr}
\end{table}

\subsection{Communication Complexity Analysis}
In this subsection, we will conduct a theoretical analysis of the communication complexity of our framework.
\subsubsection{Clustering Election Phase}
We assume that $\epsilon_1$ is the communication traffic for one heartbeat interaction between vehicles, $m$ denotes the number of vehicles, and $ p_1$ represents the probability of each vehicle participating in the election.
During the decision to participate in the election phase, an average of $mp_1$ nodes will become Candidate, and during the Candidate broadcast heartbeat phase, the following traffic will be generated: 

$$m(1-p_1)\cdot{mp_1}\cdot{\epsilon_1}$$

During the Follower voting phase, every Follower will send heartbeat signals of voting to Candidates, and the following traffic will be generated: $$m(1-p_1)\cdot{mp_1}\cdot{\epsilon_1}$$

During the counting phase, Since every two Candidates interact with each other, then the following traffic analysis results can be derived from the combination principle: 
$$\dbinom{2}{mp_1}\cdot{\epsilon_1}\cdot2$$

In the dynamic clustering process, we take the data distribution as a criterion; We assume that $\epsilon_2$ is the communication traffic for calculating the data distribution difference between vehicles. In the best case, i.e., all vehicles have similar data distribution and only one cluster is generated, the communication traffic is represented by the following equation:

$$(m-1)\cdot{\epsilon_2}$$
In the worst case, where each vehicle is divided into a separate cluster, the above equation will become as follows: 

$$\epsilon_2+2\epsilon_2+\cdots+{(m-1)\epsilon_2}=\frac{m(m-1)\cdot{\epsilon_2}}{2}$$

In the global central node election process, we adopt the scheme of sharing a small amount of data; we assume that $\widetilde T$ is the communication traffic for transmitting data between vehicles. The election process is the same as described above, and the traffic can also be expressed as follows: 

$$\widetilde T+m(1-p_1)\cdot{mp_1}\cdot{\epsilon_1}\cdot{2}+\dbinom{2}{mp_1}\cdot{\epsilon_1}\cdot2$$

\subsubsection{Model Training Phase}
We assume that $\widetilde P$ is the communication traffic of one model parameter transmission, and $M$ is the number of clusters; we evaluate on the premise that the number of every cluster is $C=\{c_1, \cdots, c_M\}$. In this case, the overall traffic for $epoch=E$ is as follows: 

$$[E\cdot(\sum_{i=1}^{M}{c_i}+M)]\cdot\widetilde P$$
For the conventional FL framework, the overall traffic for $epoch=E$ is as follows: 
$$[E\cdot(2\cdot\sum_{i=1}^{M}{c_i}+2\cdot M)]\cdot\widetilde P$$
As such, RaftFed effectively reduces communication volumes.

\section{conclusion}\label{CO}
In this paper, we proposed RaftFed, a lightweight federated learning framework suitable for VCI networks to enable privacy-preserving vehicular collaboration. Specifically, we based dynamic clustering and node selection mechanisms to implement a two-layer iterative federated learning scheme to eliminate communication bottlenecks caused by interacting with a remote server. Moreover, we incorporated a pre-training mechanism by exploiting non-sensitive data to alleviate Non-IID sufferings and accelerate the model convergence rate. Through thorough experimental evaluations, it is shown that RaftFed outperforms baselines regarding convergence rate, model accuracy, and communication overhead.

%In the future, we will further study how to enhance RaftFed robustness and security. With respect to robustness, we plan to investigate what mechanisms make RaftFed adapt to dynamic VCI networks, i.e., vehicular nodes are allowed to leave or join the network arbitrarily. Pertaining to security, we plan to research how a trusted execution environment is seamlessly integrated into RaftFed protocols and whether decentralization will contribute to a better RaftFed in terms of security and performance. Moreover, we plan to implement and deploy the RaftFed prototype in the real world.

\bibliographystyle{IEEEtran}
\bibliography{ref}
\end{document}